\documentclass{egpubl}
\usepackage{pg2020}
 
 \WsPaper           % uncomment for final version of EG Workshop contribution
\usepackage[T1]{fontenc}
\usepackage{dfadobe}  
\usepackage{amsmath}
\usepackage{amsfonts}
\usepackage{makecell}
\usepackage{cite}  % comment out for biblatex with backend=biber
\BibtexOrBiblatex
\electronicVersion
\PrintedOrElectronic
\ifpdf \usepackage[pdftex]{graphicx} \pdfcompresslevel=9
\else \usepackage[dvips]{graphicx} \fi

\usepackage{egweblnk} 
\title[A deep learning based interactive sketching system for fashion images design]%
      {A Deep Learning Based Interactive Sketching System \\ for Fashion Images Design}

\author[Yao Li \& Xianggang Yu ]
{\parbox{\textwidth}{\centering Y. Li$^{1}$, 
X.\,G. Yu$^{2}$, 
X.\,G. Han\thanks{Corresponding author: Xiaoguang Han, hanxiaoguang@cuhk.edu.cn}$^{2}$, 
N.\,J. Jiang$^{3}$, 
K. Jia\thanks{Kui Jia is with the School of Electronic and Information Engineering, South China University of Technology, Guangzhou, China, the Pazhou Lab, Guangzhou, 510335, China, and the Peng Cheng Lab, Shenzhen 518005, China}$^{4}$ 
and J. B. Lu$^{3}$
        }
        \\
{\parbox{\textwidth}{\centering $^1$DexForce Technology Co. Ltd. \quad 
         $^2$Chinese University of Hong Kong ~(Shenzhen)\\
         $^3$SmartMore Corporation \quad
         $^4$South China University of Technology
       }
}
}
\begin{document}\sloppy

\teaser{
%\begin{figure*}
  \includegraphics[width=0.85\linewidth]{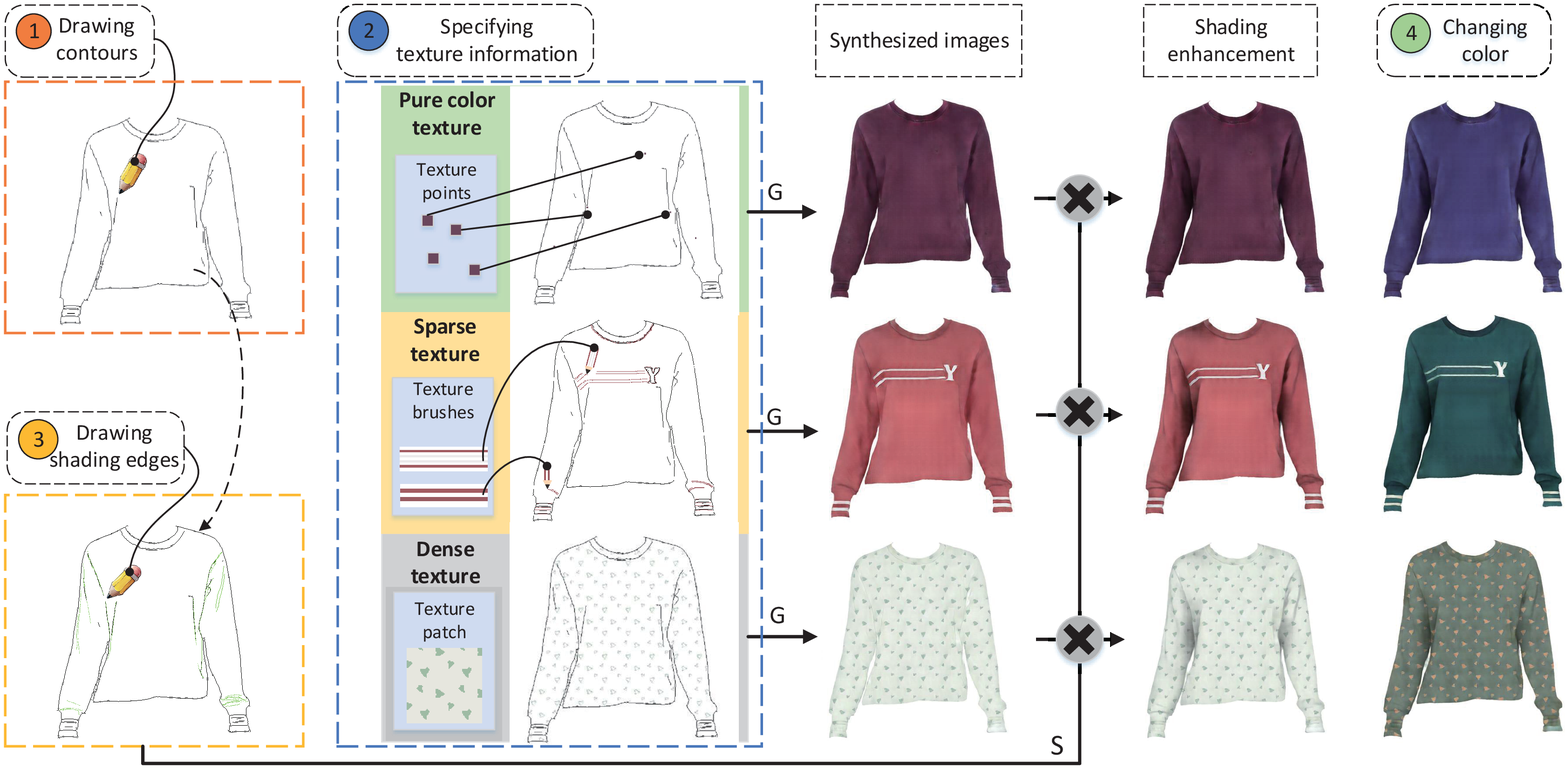}
  \centering
   \caption{The interactive pipeline to use our sketching system for fashion images design. 
   The user can firstly draw contours to determine the shape of the desired garment. Three different modes are then provided for specifying different textures on the garment, i.e., pure colored, sparse textured and dense textured. 
A generative adversarial network is adopted here to convert these users' inputs to realistic photos. 
   Thirdly, some shading edges are allowed to be drawn which will also be converted into shading information with deep learning methods, making the garment image enhanced to be more stereoscopic. As post-processing, all the colors can be further modified straightforwardly. 
}
 \label{fig:interaction_system}
% \end{figure*}
}

\maketitle
%-------------------------------------------------------------------------
\begin{abstract}
   In this work, we propose an interactive system to design diverse high-quality garment images from fashion sketches and the texture information. The major challenge behind this system is to generate high-quality and detailed texture according to the user-provided texture information.
   Prior works mainly use the texture patch representation and try to map a small texture patch to a whole garment image, hence unable to generate high-quality details. In contrast, inspired by intrinsic image decomposition, we decompose this task into texture synthesis and shading enhancement. In particular,  we propose a novel bi-colored edge texture representation to synthesize textured garment images and a shading enhancer to render shading based on the grayscale edges. 
   The bi-colored edge representation provides simple but effective texture cues and color constraints, so that the details can be better reconstructed. Moreover, with the rendered shading, the synthesized garment image becomes more vivid.
%  Extensive experiments on a manually-collected fashion garment dataset demonstrate that our system can generate various high-resolution and photorealistic garment images. 
%-------------------------------------------------------------------------
%  ACM CCS 1998
%  (see https://www.acm.org/publications/computing-classification-system/1998)
% \begin{classification} % according to https://www.acm.org/publications/computing-classification-system/1998
% \CCScat{Computer Graphics}{I.3.3}{Picture/Image Generation}{Line and curve generation}
% \end{classification}
%-------------------------------------------------------------------------
%  ACM CCS 2012
%   (see https://www.acm.org/publications/class-2012)
%The tool at \url{http://dl.acm.org/ccs.cfm} can be used to generate
% CCS codes.
%Example:


\end{abstract}  

%-------------------------------------------------------------------------
\section{Introduction}

\label{sec:intro}
In recent years, a growing number of customers are looking for a more exclusive personalized customizing, especially garment customization. The key technique of garment customization is to synthesize a corresponding cloth image based on user-provided design elements like contours, fabric patterns and so on.
Existing garment-drawing software often requires professional experiences and the drawing process is also time-consuming. It is arduous for a novice user to design a garment using common software. Therefore, it is meaningful to design a user-friendly interactive tool to meet users' demands while alleviating their workloads.
 With the invention of conditional generative adversarial networks~(cGANs) in~\cite{isola2017image}, we can formulate garment-drawing as an image-to-image translation problem, which can be solved by the Pix2pix~\cite{isola2017image,wang2018high} framework. 
% Unlike unconditional GANs, the inputs of the discriminator of conditional GANs not only include images from a training set or generated by the generator, but also concatenate conditions such as discrete labels, and edge maps, which offer more control over the generated image.

Based on Pix2pix~\cite{isola2017image,wang2018high} framework, a few methods have been proposed for garment image generation. Xian \textit{et al}.~\cite{xian2018texturegan} introduced a textureGAN framework to generate garment images from an input contour map and a user-provided texture patch. They propose a new loss function to guide the network to propagate the texture patch to the corresponding regions of the given contour map, but its results require the correct position and proper size of the texture patch. To generate a variety of texture pattern outputs, similar to BicycleGAN~\cite{zhu2017unpaired}, FashionGAN~\cite{cui2018fashiongan} encodes the color and material information into a latent vector through an encoder network, rather than directly taking texture patch as the input. However, prior methods suffer from two drawbacks: 1) only a small texture patch is propagated to a whole garment image under the constraint of the contour map. Although this input setting requires less efforts for users, it is difficult to generate a high-quality garment image containing wrinkles and shading solely from a small-size texture pattern; 2) the contour map with the texture patch representation are not sufficient to generate fashion garment with more designing details, e.g. garment decoration and logos. An intuitive solution for this is to represent the detailed texture patterns with binary edges, and combined with several color points to synthesize a garment image. But as Dekel \textit{et al}.~\cite{dekel2018sparse} demonstrated it is insufficient to generate a high quality image from a binary edge map. Moreover, we empirically find that this method often leads to blurry results at the edge areas.

\begin{figure}
\centering
%\vspace{-0.15in}
\includegraphics[width=0.7\columnwidth]{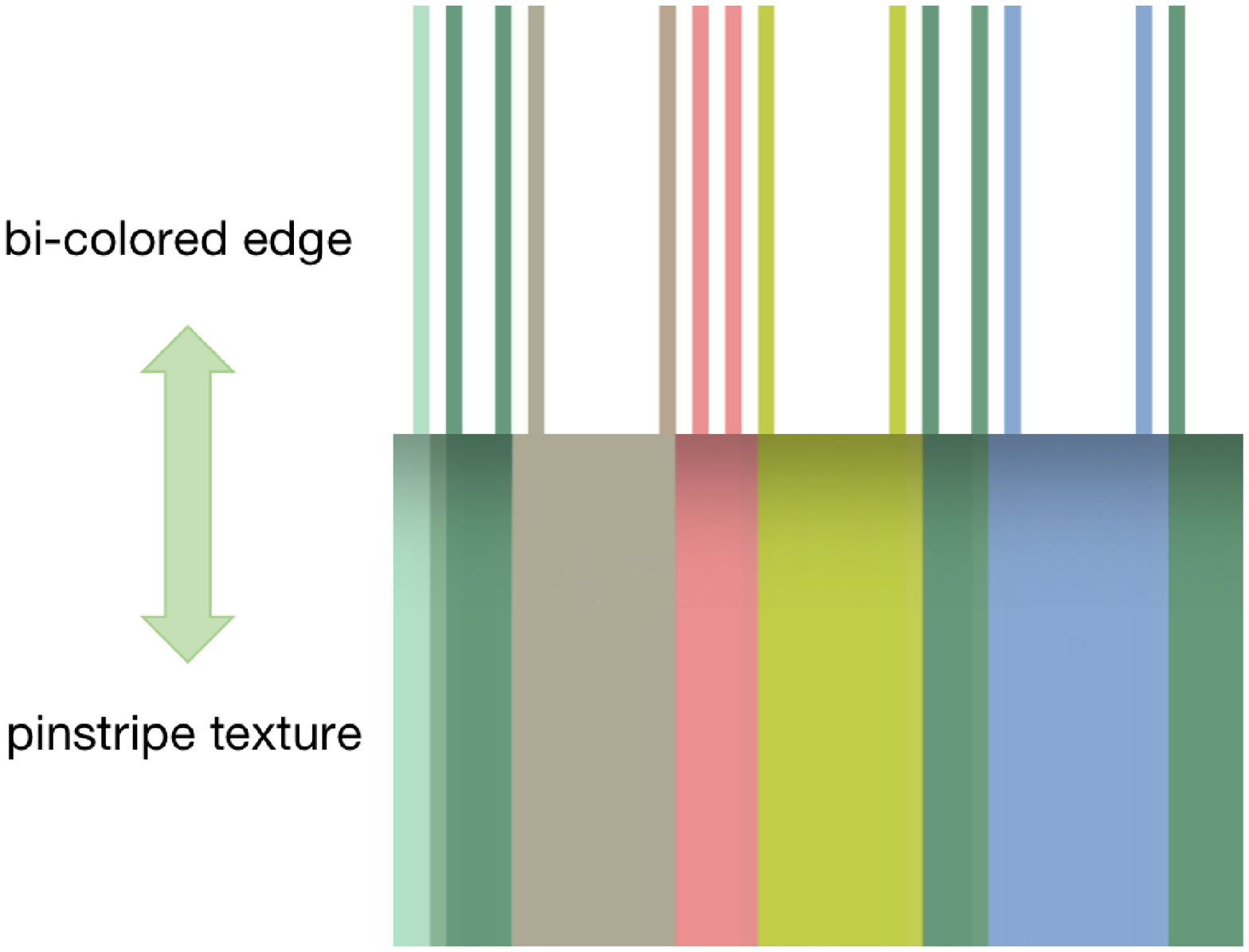}
\caption{An illustration of the bi-colored edge representation. We extract the color point at both sides of the texture edges as our texture information.
}
\label{fig:data}
\end{figure}

Therefore, based on the above discussions, we decompose this task into texture synthesis and shading enhancement. 
%which makes a trade-off between the quality of the synthesized image and the demand of user's effort.
Particularly, we propose a bi-colored edge texture representation, that is, extract the color points at both sides of texture edges as our texture information to synthesize textured garment images, as shown in Figure.~\ref{fig:data}. The bi-colored edge representation provides simple but effective texture cues and color constraints, so that the details can be better reconstructed. We also introduce a shading enhancer to render shading on a garment image by simply drawing some sparse binary edges. In this way, the user can achieve more control on the shading information of a garment image.
Even though the bi-colored edge representation will increase the demand of users' interactions compared with the patch representation, our experiments demonstrate this representation has sufficient capability to generate diverse high-quality garment images.
%Some reconstruction results of our method are shown in Figure~\ref{fig:examples}. 
We also design a handy interactive system based on this representation. It allows free-form design such as garment decoration, logo and etc.

%Finally, in order to manipulate the synthesized color during inference, we utilize a color constraint based on KL loss function to restrain the color distribution of the output, which can guide the network to propagate the input color distribution to the output. This loss designing makes the synthesized RGB image match arbitrary user-provided colors, rather than limited on those appeared in the training set.

Our major contributions in this work can be summarized as follows:
\begin{itemize}
  \item We propose an elaborate interactive-system for fashion images design. It is convenient for users to customize personalized fashion garments.
  \item We decompose the task of garment images synthesis into texture synthesis and shading enhancement, and propose the bi-colored edge representation which reaches a good balance between the quality of the synthesized images and the workload of users.
  \item We propose a shading enhancer for interactively drawing grayscale edges and separately inferring the shading information. This greatly facilitates the photo-realism of the results.
\end{itemize}

%------------------------------------------------------------

%-------------------------------------------------------------------------
\section{Methodology }
\label{sec:method}

%We decompose the task of garment image synthesis into texture synthesis and shading enhancement. 
Our framework includes two generators, namely garment generator and shading generator. 
The former generates color images based on the fashion contours and the proposed bi-colored edge representation. The latter one renders shading based on the shading edges.
% in an image decomposition way.
And then an interactive system is proposed for generating high-quality fashion garment images based on the aforementioned two generators as shown in Figure~\ref{fig:interaction_system} . 

%-------------------------------------------------------------------------
\subsection{Image synthesis from sketches}
\label{sec:model}
\begin{figure}
\centering
%\vspace{-0.15in}
\includegraphics[width=0.9\columnwidth]{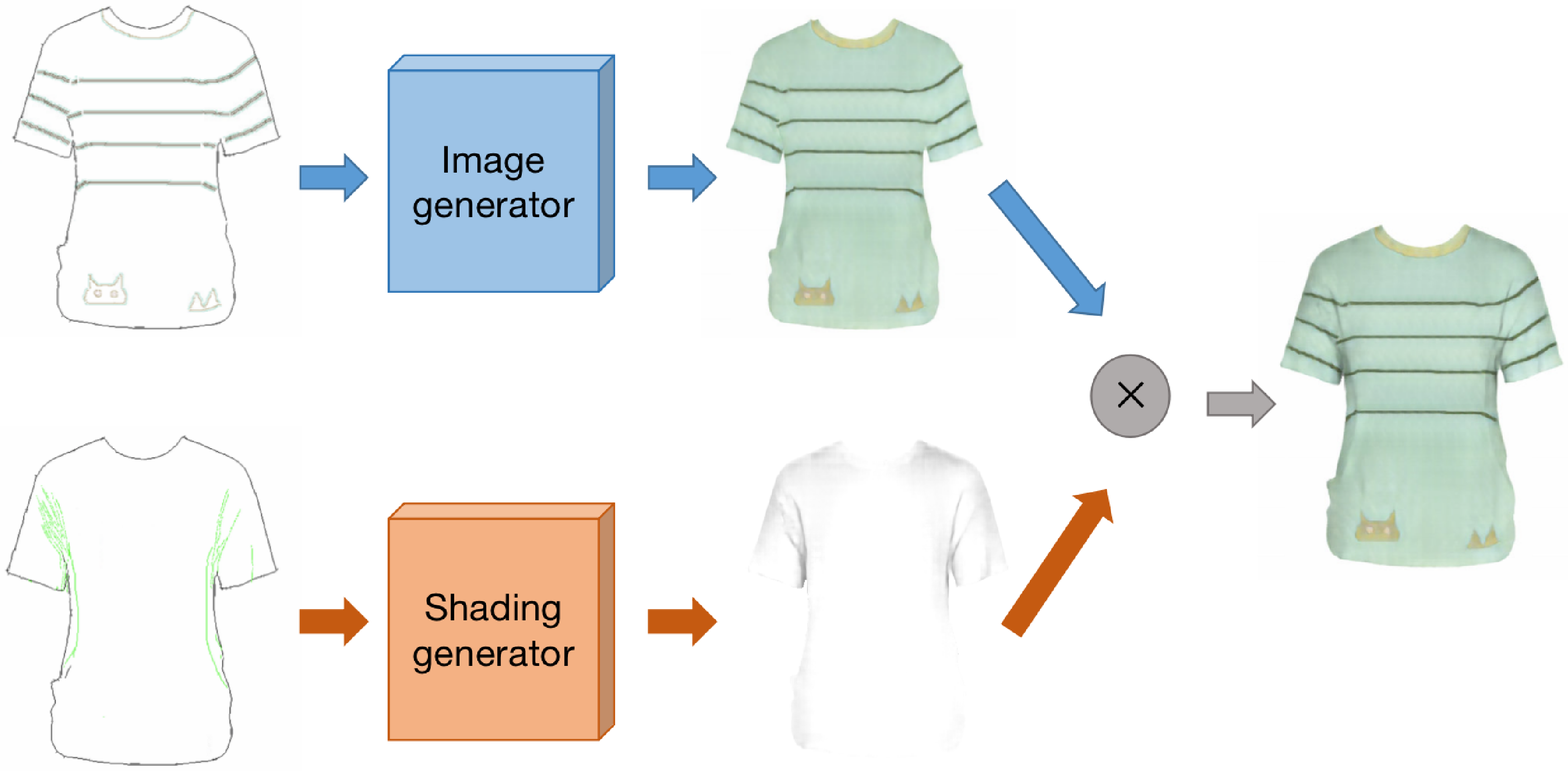}
\caption{The framework contains an image generator and a shading generator. The former generates color images based on the fashion contours and the proposed bi-colored edge representation. Then the latter renders
shading based on the shading edges in an image decomposition way.
}
\label{fig:pipeline}
\end{figure}

In this subsection, we first introduce how to train an image generator to synthesize high-resolution, colored clothing image, and train a shading generator to improve the stereoscopic shape of the synthesized garment image. The framework is shown in Figure~\ref{fig:pipeline}

%%-------------------------------------------------------------------------

%-------------------------------------------------------------------------
\subsubsection{Fashion image generator}
Our fashion image generator network is based on the architecture proposed by Johnson \textit{et al}.~\cite{johnson2016perceptual}, which is a deep residual convolutional neural network. It consists of three components: a convolutional layer to downsample the input, a set of residual blocks~\cite{he2016deep, gross2016training} and a convolutional layer to upsample. The network downsampling and then upsampling allow the network to have an effective receptive field size on the input, making the generated image more realistic. 
%Figure~\ref{fig:network} shows our network architecture in detail.
The input of our fashion image generator is a tensor of size $512 \times 512 \times 4$: a binary sketch that describes the structure of a garment image, a RGB bi-colored edges map.

To enhance the performance of the generator, we adopt the framework of cGAN~\cite{isola2017image} which has been used intensively for image-to-image translation tasks. Thus both the contour map and the bi-colored edges are inputted into the discriminator as a condition to guide the generator to synthesize more realistic image. In addition, to synthesize high-resolution garment images, we adopt a multi-scale discriminator framework similar to Wang \textit{et al}.~\cite{wang2018high}. 
% As shown in Figure~\ref{fig:network}, we downsample the conditional generator input, the synthetic garment image or the target garment image by a factor of 2 and used as the input of the second discriminator.
Those discriminators are trained to distinguish real and synthesized image at different scales, respectively. At each scale, the discriminator has different receptive fields on the input. 

In our experiment, we apply lsGAN~\cite{mao2017least} to train the generator and the discriminator.
The adversarial loss of the discriminator is written as:
%-------------------------------------------------------
\begin{equation}
\begin{aligned}
  \mathcal L_{adv}(\mathit{D}) = &\sum_{s=1}^S \frac{1}{2} \mathbb E_{(\mathbf {x,y}){\sim}p_{data}(\mathbf {x,y})}{\left[{\bigl(D_s(\mathbf {x,y})- 1\bigr)^2}\right]}\\
 &+ \frac{1}{2} \mathbb E_{\mathbf x{\sim}p_{data}(\mathbf x)}{\left[{D_s(\mathbf  x, G(\mathbf x))^2}\right]} ,
\end{aligned}
\end{equation}
where the $S$ is the number of discriminators in the network architecture.
%-------------------------------------------------------
Also, the adversarial loss of generator is:
\begin{equation}
\mathcal L_{adv}\mathit{(G)} = \mathbb E_{\mathbf x{\sim}p_{data}(\mathbf x)}{\left[{\bigl(D_S(\mathbf  x, G(\mathbf x)) - 1\bigr)^2}\right]} .
\end{equation}

%-------------------------------------------------------
$L1$ loss is considered to ensure the generated image $\mathit{G(\mathbf x)}$ from the input $\mathbf{x}$ to be close to its ground truth image $\mathbf{y}$. It is given by:
\begin{equation}
  \mathcal L_{L1}\mathit{(G)} = \mathbb E_{(\mathbf {x,y})}{\sim}p_{data(\mathbf {x,y})}{\left[{\| {\mathbf y - G(\mathbf x)}\|}_1\right]} .
\end{equation}

%-------------------------------------------------------
We also adopt a perception loss to increase the quality of the generated image. The perceptual loss measures high-level perceptual and semantic differences between two images by projecting images into feature spaces using VGG-19~\cite{simonyan2014very} which is pre-trained on the ImageNet. 
\begin{equation}
  \mathcal L_{p}\mathit{(G)} = \sum_{i=1}^{\mathit N} \frac{1}{\mathit{M_i}} {\left[{\| {\Theta^{(i)}(\mathbf y) - \Theta^{(i)}(G(\mathbf x))}\| }_1\right]} .
\end{equation}
Here, $\Theta^{(i)}(\cdot)$ is the feature map with $\mathit{M_i}$ elements, as the output from the $i$-th layer of VGG network.

%-------------------------------------------------------
Inspired by the color unmixing algorithm~\cite{aksoy2016interactive},  we introduced a KL loss function to restrain the color distribution of the output to be consistent with the input's.
The color constraint loss function is defined as:
\begin{equation}
\begin{aligned}
  \mathcal L_{kl}\mathit{(G)} 
 = &\sum_{i=1}^{\mathit K} \frac{1}{2} \Bigl( {log\frac{|{\Sigma_{\tilde{\mathbf y}}^{(i)}}|}{|{\Sigma_{\mathbf y}^{(i)}}|}} - n + tr({\Sigma_{\tilde{\mathbf y}}^{(i)}}^{-1}\Sigma_{\mathbf y}^{(i)}) \Bigr.\\
&+ \Bigl. (\mu_{\tilde{\mathbf y}}^{(i)} - \mu_{\mathbf y}^{(i)})^T{\Sigma_{\tilde{\mathbf y}}^{(i)}}^{-1}(\mu_{\tilde{\mathbf y}}^{(i)} - \mu_{\mathbf y}^{(i)}) \Bigr) .
\end{aligned}
\label{eq:kl}
\end{equation}
Here, we denote the output of the generator $G(\mathbf x)$ as $\tilde{\mathbf y}$. We use the hierarchical clustering algorithm to obtain a set of color clusters and $\mathit K$ is the number of color cluster. $\mu^{(i)}$ and $\Sigma^{(i)}$ are the mean vector and the covariance matrix of the $i$-th color cluster calculated by using the cluster label. 

%-------------------------------------------------------
The total generator loss function is written as:
\begin{equation}
\begin{aligned}
\mathcal L = \lambda_{adv}\mathcal L_{adv} + \lambda_{L1}\mathcal L_{L1} + \lambda_{p}\mathcal L_{p} + \lambda_{kl}\mathcal L_{kl} .
\end{aligned}
\label{eq:garment}
\end{equation}

%-------------------------------------------------------
\subsubsection{Shading enhancement}
\label{sec:shading}
The image generator can synthesize detailed texture patterns, but the results are likely to lack shading information.
Thus, we introduce a shading generator to synthesize shading information from a contour map and sparse binary edges. Particularly, we are not directly to synthesize the shading-corrected image, but to synthesize a shading image which is a single channel image.
We use the notion of intrinsic image decomposition~\cite{bi20151, li2018learning} to obtain the shading image. That is, a RGB image $I \in \mathcal R^{H\times W\times 3}$ can be factorized into a product of a reflectance image $R \in \mathcal R^{H\times W\times 3}$ and a shading image $S \in \mathcal R^{H\times W}$, $I=R \times S$. With that, the synthesized garment image by fashion image generator is regarded as $\tilde R$, the result of the shading generator is $\tilde S$, and the final result of our system is $\tilde{S}\times \tilde R$.
The network architecture is the same as the garment generator, except that we change the last activation layer from Tanh function to ReLU. Since the pixel value range of the shading image is relatively discrete, it is unwise to conduct zero-centered operation on it.

The loss function to train the shading generator is shown as below:
\begin{equation}
\mathcal L = \lambda_{rec}\mathcal L_{rec} + \lambda_{dense}\mathcal L_{dense} .
\label{eq:shading}
\end{equation}
The reconstruction loss $\mathcal L_{rec}$ computes the distance between the ground truth image $I=S\times R$ and the synthesized image $\tilde{S}\times R$ in RGB domain. 
\begin{equation}
  \mathcal L_{rec} = \mathbb E_{(\mathbf {S,I})}{\sim}p_{data(\mathbf {S,I})}{\left[{\| { I - \tilde{S}\times R }\|}_1\right]} .
\end{equation}

As the shading information of a garment image is sparse,  and the final result $\tilde{S}\times \tilde R$ should be similar with $\tilde R$, that is most of the pixels value in synthesized shading image $\tilde S$ should close to 1. Thus we conduct the dense loss:
\begin{equation}
  \mathcal L_{dense} = \mathbb E_{(\mathbf {S,I})}{\sim}p_{data( \mathbf {S,I})}{\left[{\| {\tilde{S} - 1}\|}_1\right]} .
\end{equation}

%-------------------------------------------------------------------------
\subsection{Dataset construction}
\label{sec:dataProcessing}
\subsubsection{Clothing images collection}
In this work, we focus on high-resolution garment fashion design. The garment image should have pure background colours as well as diverse fashion styles. Although the public large-scale garment dataset like DeepFahion~\cite{liuLQWTcvpr16DeepFashion} contains a large number of garment images including various textures, most of the clothing styles are T-shirt and lack rich fashion styles. 
We therefore collect a fashion garment dataset which have clear background, high-resolution images as well as diverse styles. The dataset contains top clothing images, skirts and coats around 4,300, of which 3900 are used as the training set and the remaining are used as the validation set. We set aside the pants category as their styles are usually relatively simple. The resolution of image in our dataset is 512$\times$512.

%------------------------------------------------------------
\subsubsection{Training samples preparing}
To train the garment generator and the shading generator, the training samples include a contour map, texture information and shading edges. 
The contour map is applied in image generator as well as shading generator. 
We apply the commonly used edge detector, HED dege detector~\cite{xie2015holistically}, to obtain edges as our contour map and then simplify those edges with an additional post-processing method provided by pix2pix~\cite{isola2017image}. 
\paragraph*{} 
For image generator, the texture edge information is obtained by extracting the color information at both sides of the texture edges.
Specifically, we first extract the texture edges by Canny edge detector~\cite{canny1987computational} and remove the outermost edges to make sure the shape of the generated image is constrained by the contour map.
% to get rid of the contour map which makes sure the shape of the generated image is constrained by the contour map.
Taking the convenience of interaction into account, we also remove the corner points.
Then the texture edge information is extracted by computing the orientation of the texture edges and sample the color points at both sides of a texture edge for each edge pixel.
% A specific example is shown in Figure~\ref{fig:data}. 
An observation is that the extracted texture edge information is in the form of double-lines, thus we can draw the information by using various types of brush during the test phase as shown in Figure~\ref{fig:interaction_system}.

\paragraph*{} 
For shading generator, we need a pair of training sample~(shading edge, shading image). Based on the notion of intrinsic image decomposition, We select about 2000 garment images which include large areas of pure colors, and compute the mean RGB value of the largest area and extend it to corresponding areas as the reflectance image $R$. The shading image is obtained by $S=I/R$.
The texture edges extracted by Canny detector is regarded as our shading edges, and we only retain the largest area of a garment image by means of color clusters.

%------------------------------------------------------------
\begin{figure}
  \centering
  \includegraphics[width=.9\linewidth]{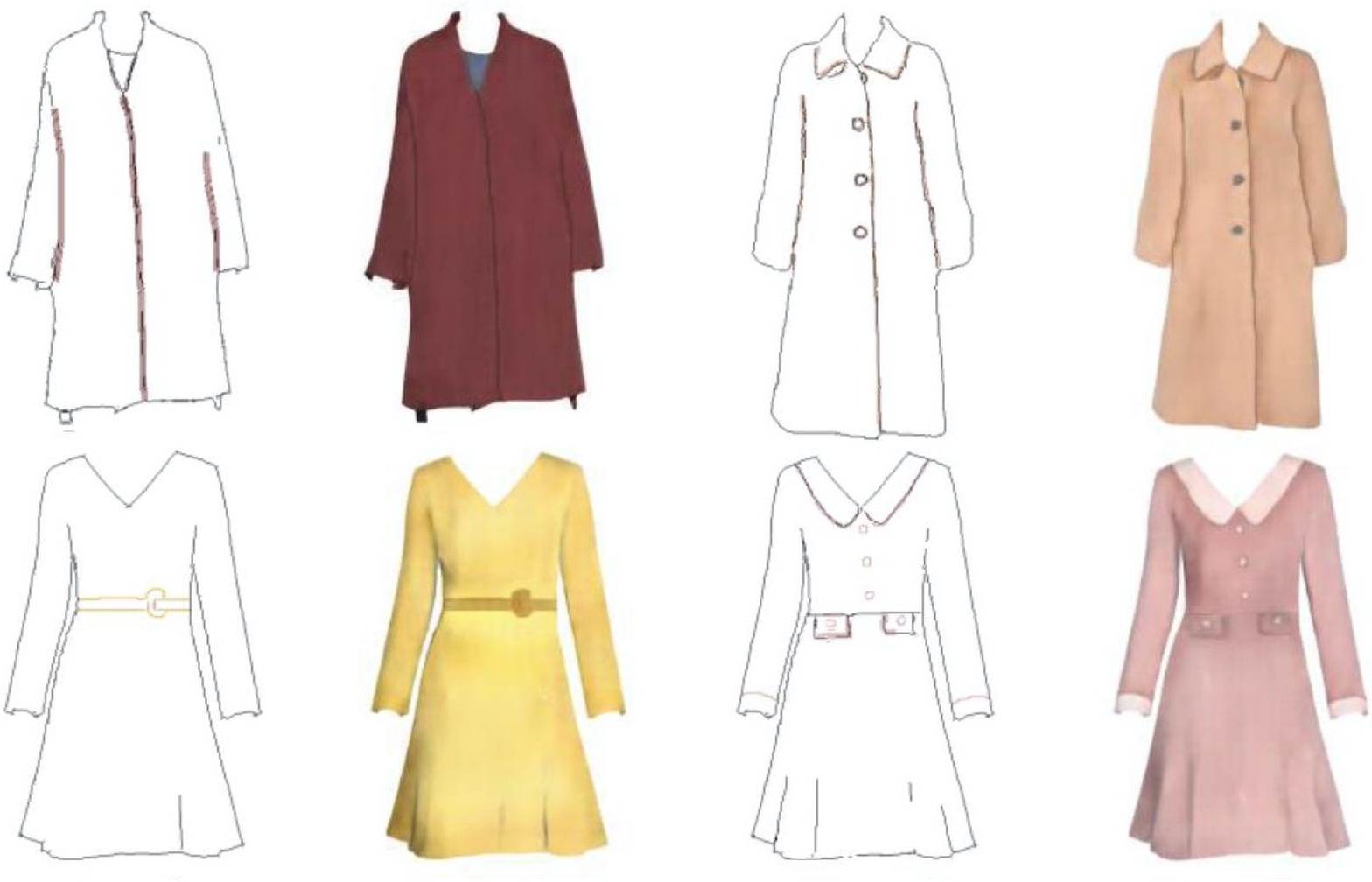}
  \caption{\label{fig:interaction_sample}%
           Those interaction samples are free form designed by using various texture brush. With the representation of texture edges, free form design such as garment decoration, logo is allowed. }
\end{figure}

%-------------------------------------------------------------------------
\subsection{User interface}
\label{sec:inter}
As illustrated in Figure~\ref{fig:interaction_system}, the UI items include a contour map, texture information and an optional item, shading edges.

\paragraph*{Drawing contours}
The first step is to determine the overall shape of a fashion garment. This can be achieved by obtaining a contour map either from user's interaction or automatically extracted from a garment image by our system. For the former input setting, a simple outline sketch is sufficient. Moreover, since the contour map is grayscale, it is easy for user to draw or manipulate them in our interactive system.

\paragraph*{Specifying texture}
For allowing users to better provide texture information in our interactive system, we design three kinds of interactive tools to meet diverse demands. 
We also classify texture pattern into three categories based on how difficult to draw them, which include pure color texture pattern, sparse texture pattern and dense texture pattern. 
For pure color texture pattern, the user only need to provide a few color points to control the color of the synthesized garment image. 
For sparse texture pattern, such as stripes and garment decoration, our system supports using a variety of brushes to offer diverse styles of design. Those brushes are used to draw the bi-colored edges, and the colors of both sides are selected by a user.
We illustrate two types of sparse texture pattern~(2 strings and 4 strings) in Figure~\ref{fig:interaction_system}. The first type is used to draw a variety of texture shapes and the second type is able to draw the pinstripe texture. 
For dense texture pattern, such as floral and tartan which often include recurrent patterns, the user is required to provide a small texture patch, and our interactive system leverages the patch match algorithm to obtain a complete texture edges by enlarging this patch to any size we want, and then extract the bi-colored edges on it. This is different from TextureGAN~\cite{xian2018texturegan} and FashionGAN~\cite{cui2018fashiongan} whose intention are to implicitly learn the PatchMatch algorithm by using the neural network.
\paragraph*{Adding shading}
The user can also augment shading on the synthesized garment image by drawing grayscale shading edges. The shading information is garment-dependent, as different shapes have different shading. 
Thus it should be used cooperatively with the contour map to synthesize factual shading information, and it should be occurred in the proper area, such as the armhole of a T-shirt or the hem of a dress. The shading edges may overlap with contour edges, but it is not contradictory since the contour edges is used to synthesize the shape of a garment, while the shading edges is applied to enhance the shading. 

%------------------------------------------------------------
\begin{figure}
\centering
%\vspace{-0.15in}
\includegraphics[width=0.85\linewidth]{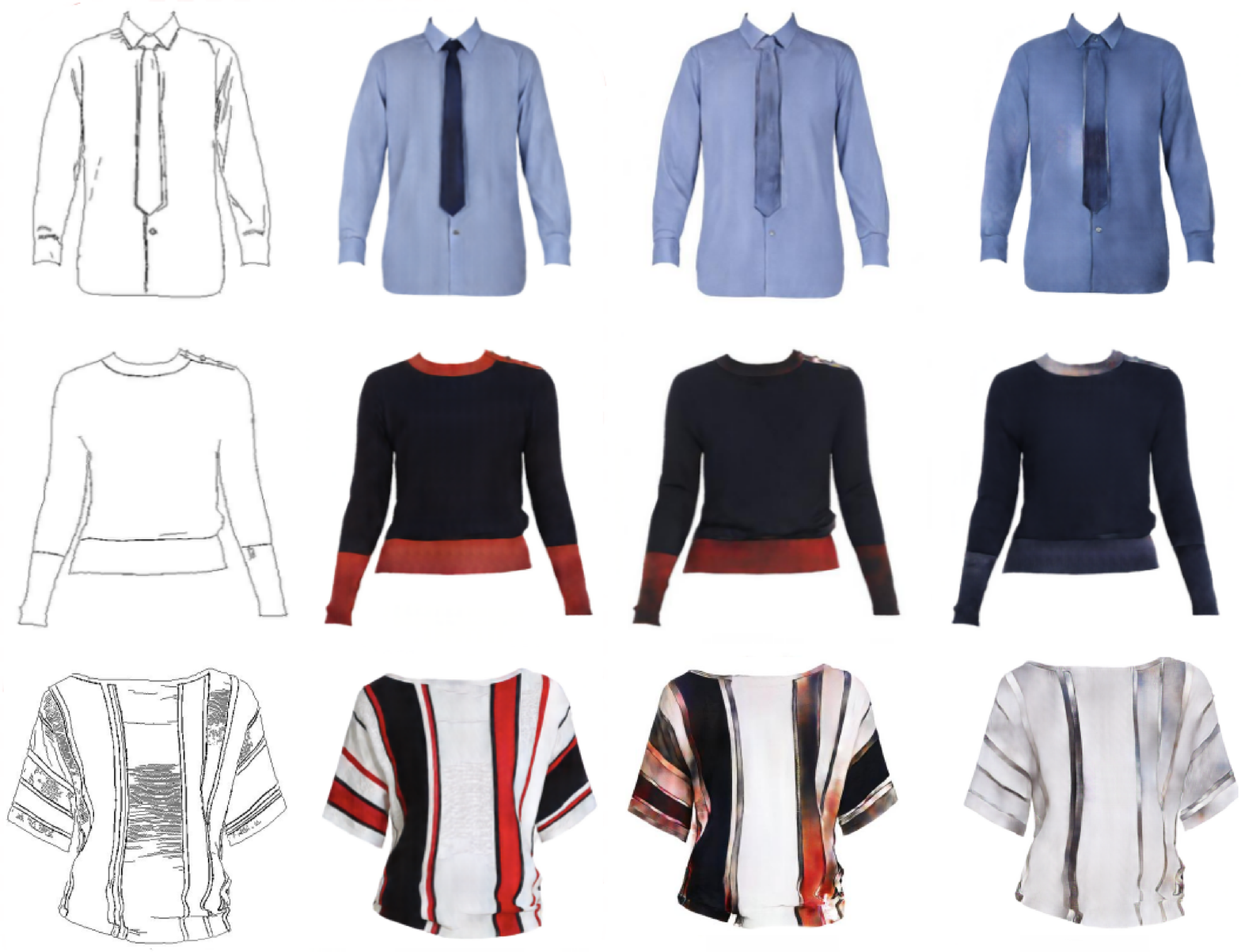}
\caption{The qualitative comparisons on the reconstruction results of different inputs. Column 2 are our results. Following columns are the results of grayscale edge and texture patch separately.  }
\label{fig:examples}
\end{figure}

\paragraph*{Changing color}
Given that the abundant color of the garments, our system also supports to change the color of the synthesized garment image. 
Specially, for pure color texture pattern, it is easy to directly change the color of the previous given color points. 
For sparse texture pattern, to avoid redraw the bi-colored edges, we can select a color used in previous stage and then replace this color with another color.
%For dense texture pattern, we can first change the color of texture image by using the K-means algorithm, in which the number of color cluster is decided on line. And then extract the bi-colored edges on it.
For a dense texture pattern, we can correctly classify the pattern's color by using the K-means algorithm, in which the number of color clusters is decided on line, and then to change the color of each cluster. Finally, we can re-extract the bi-colored edges on it and to synthesize garment images.

%The full process of interaction design is shown in Figure~\ref{fig:UITest}.
 Our system consists of three interfaces, namely drawing interface, color regulator and output interface. The drawing interface includes three layers, the garment contour layer, the bi-colored texture edge layer and the shading edge layer. The garment contour edge and shading edge are both grayscale, it is easy for user to draw or manipulate them in our interactive system. With the designed interactive tools, it is also convenient to draw bi-colored edge. To change the color of the synthesized garment image, we collect the occur color on the bi-colored edge layer and show those color in the right-top corner. We only need to change the color of this corner, and the corresponding color in bi-colored texture edge layer will be changed.
 
%-------------------------------------------------------
\section{Experiments}
\subsection{Implementation details}
To train the garment generator, we employ two discriminators to distinguish real and fake images at two image scales. Our algorithm is implemented in PyTorch and trained on two GTX 1080 GPUs for 150 epochs using the ADAM optimizer with lr = 0.0002 and beta1 = 0.5, beta2 = 0.999, consuming two days.
For the garment generator, we set different weights for the losses, where $\lambda_{adv}=1, \lambda_{L1}=10, \lambda_{p}=10, \lambda_{kl}=0.01$ in Eq.~(\ref{eq:garment}).
For the shading generator, we set $\lambda_{rec}=100, \lambda_{dense}=1$ in Eq.~(\ref{eq:shading}).

%------------------------------------------------------------
\begin{figure}
\centering
\includegraphics[width=\linewidth]{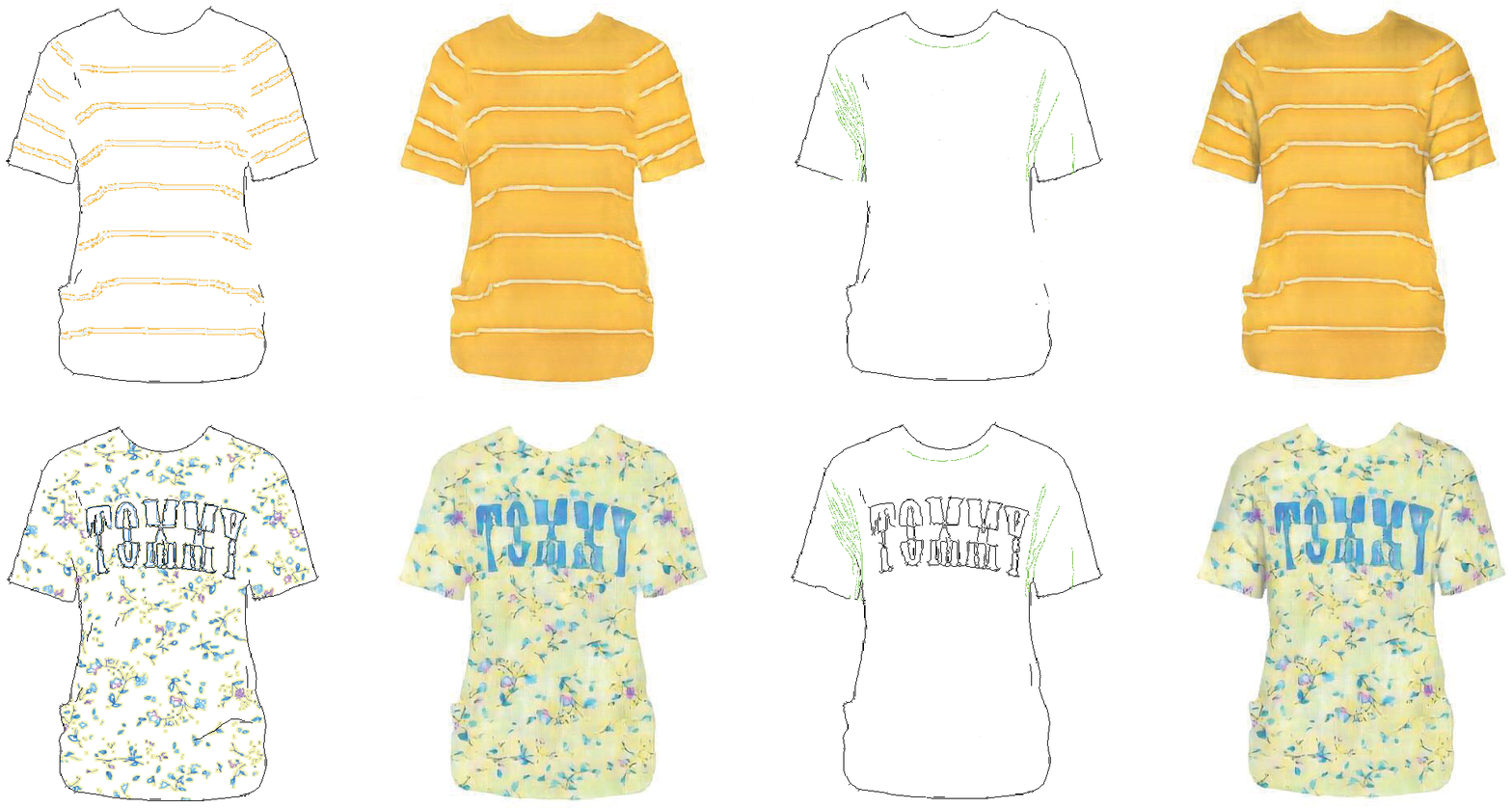}
\caption{The qualitative comparisons on shading enhancement. The second column is the result of the image generator, and the last column is the result of the shading enhancement. }
\label{fig:shadingEnhance}
\end{figure}

%-------------------------------------------------------
\subsection{ Results and analysis}
In this subsection, we firstly show more qualitative results designed by our sketching system. 
Then we also conduct quantitative and qualitative evaluations on the proposed bi-colored edge representation with some other variants.

\subsubsection{Results gallery}
As shown in Figure~\ref{fig:interaction_sample}, our interactive system can design diverse and colorful garment. With the flexibility of the bi-colored edge representation and the shading generator to increase the photo-realism of the results, we can draw various texture patterns and synthesize high-quality garment image.  
Although drawing is required, our interactive system provides very convenient drawing tools and is suitable for any synthesized resolution. 

%------------------------------------------------------------
\begin{table}
\small
\begin{center}
\begin{tabular}{ |c|c|c|c|c| }
  \hline
    & \makecell[c]{Patch} & \makecell[c]{Binary edge + \\ color points } &
  \makecell[c]{Bi-colored \\ edge} & \makecell[c]{Bi-colored \\ edge + KL } \\
  \hline
  IS & 3.655 & 3.772 & 4.074 & \textbf{4.076} \\  % Middlebury 1
  \hline
  FID & 33.220 & 33.813 & 18.460 & \textbf{18.393} \\ % Lamp Shade 2
  \hline
   User study & 1.949 & 2.062 & 2.713 & \textbf{2.717} \\ % Lamp Shade 2
  \hline
\end{tabular}
\end{center}
\caption{Quantitative comparison of the there texture representation. Our Bi-colored edge representation achieves a higher score in all metrics. 
%The KL loss~(Eq.~(\ref{eq:kl})) dose not realize large performance improvement, since it is used to restrain the consistency of color distribution rather . 
A higher IS and lower FID values means a better performance.}
 \label{table:quantitative}
\end{table}

\subsubsection{Comparisons on different user inputs }
One of the key contributions in this work is the proposed new texture representation, that is the bi-colored edge. So we compare the results of different texture representation. Compared with our bi-colored edge representation, the two other variants can be used as the inputs: 1) use binary edges to determine texture information and sampled color points to specify colors; 2) directly use a texture patch without a PatchMatch method, say the method of textureGAN~\cite{xian2018texturegan} . For training the first network, we sample a small area of size from $1 \times 1$ to $9 \times 9$ on the real image and compute the mean color of this small area to obtain the sparse color points in the training phase. The number of color points is from 50 to 100. For training the second network, a garment patch of a random size from $50 \times 50$ to $70 \times 70$ is sampled from a real garment image, and the patch location is mainly chosen in the center regions of a garment.

Figure~\ref{fig:examples} shows the reconstruction results of the three kinds user input. The groundtruth garment images are selected from our dataset which are not used during the training procedure. 
Our bi-colored edges representation can synthesize more unambiguous texture patterns compared with the input of the binary edges and several color points. Although a local texture loss is considered~\cite{xian2018texturegan}, the texture patch representation based method is still hard to synthesize complex texture patterns. On the contrary, our bi-colored edges representation can synthesize high-quality garment image and accord with the texture pattern given by user. 

We then use Inception Score~(IS) and Frechet Inception Distance~(FID)~\cite{heusel2017gans} evaluation metrics which are widely used in generative models to evaluate the performance of the two different representations. We also conduct a user study to assess the realness of the generated images and compare the performance of our proposed bi-colored edge representation with two other texture representations. 10 participants are shown 50 synthesized image pairs, and then grade each pair images in authenticity. The highest score is 3 and the lowest is 1.  Table~\ref{table:quantitative} proves that our bi-colored edge representation is valid. And our proposed color constraint loss function is improved in both evaluation metrics, which indicates that it can increase the diversity and quality of the generated images.

\subsubsection{Ablation study on shading enhancement}
Figure~\ref{fig:shadingEnhance} shows the qualitative evaluations about our shading enhancement. It is obviously that the shading enhancement garment images become more vivid. We also performed a user study for evaluating the visual quality of shading enhancement. 10 participants were shown 30 garment image pairs with and without shading enhancement, and then are asked to which image is more photo-realism. It will be scored if the participant answers correctly within a second, otherwise it will not be scored. The final average score is 26/30. This result also demonstrates that our shading enhancer can effectively render the shading on a garment image, so as to improve the realism of results.
%-------------------------------------------------------------------------
\section{Conclusion}
In this work, we decomposed the task of garment images synthesis into texture synthesis and shading enhancement and proposed an interactive sketching system for fashion image synthesis. The system allows users to draw contours, specifying diverse texture information and performing shading enhancement by sketching shading edges. Deep learning based methods are applied to convert these sparse information to realistic photos. To well balance the output quality and interaction workload, a novel bi-colored edge representation is proposed. Experiments demonstrate the superiority of such representation against all others. 

\section{Acknowledgements}
The work was supported in part by the Key Area R$\&$D Program of Guangdong Province with grant No. 2018B030338001, by the National Key R$\&$D Program of China with grant No. 2018YFB1800800, by Natural Science Foundation of China with grant NSFC-61902334 and NSFC-61629101, by Guangdong Zhujiang Project No. 2017ZT07X152, and by Shenzhen Key Lab Fund No. ZDSYS201707251409055, by the Program for Guangdong Introducing Innovative and Enterpreneurial Teams (Grant No.: 2017ZT07X183), the National Natural Science Foundation of China (Grant No.: 61771201), and the Guangdong R$\&$D key project of China (Grant No. 2019B010155001), by the Shenzhen Cloudream Technology Co. Ltd.\footnote{This work was mainly done when Yao, Nianjuan and Jiangbo were interning and working in Cloudream Technology.}

%-------------------------------------------------------------------------
% bibtex
\bibliographystyle{eg-alpha-doi} 
\bibliography{egbibsample}       

\newcommand{\etalchar}[1]{$^{#1}$}
\begin{thebibliography}{\uppercase{DGK{\etalchar{*}}18}}

\bibitem[AAPS16]{aksoy2016interactive}
\textsc{Aksoy Y., Aydin T.~O., Pollefeys M., Smoli{\'c} A.}:
\newblock Interactive high-quality green-screen keying via color unmixing.
\newblock \emph{ACM Transactions on Graphics (TOG) 36}, 4 (2016), 1.

\bibitem[BHY15]{bi20151}
\textsc{Bi S., Han X., Yu Y.}:
\newblock An l 1 image transform for edge-preserving smoothing and scene-level
  intrinsic decomposition.
\newblock \emph{ACM Transactions on Graphics (TOG) 34}, 4 (2015), 78.

\bibitem[BSFG09]{barnes2009patchmatch}
\textsc{Barnes C., Shechtman E., Finkelstein A., Goldman D.~B.}:
\newblock Patchmatch: A randomized correspondence algorithm for structural
  image editing.
\newblock In \emph{ACM Transactions on Graphics (ToG)} (2009), vol.~28, ACM,
  p.~24.

\bibitem[Can87]{canny1987computational}
\textsc{Canny J.}:
\newblock A computational approach to edge detection.
\newblock In \emph{Readings in computer vision}. Elsevier, 1987, pp.~184--203.

\bibitem[CLGS18]{cui2018fashiongan}
\textsc{Cui Y.~R., Liu Q., Gao C.~Y., Su Z.}:
\newblock Fashiongan: Display your fashion design using conditional generative
  adversarial nets.
\newblock In \emph{Computer Graphics Forum} (2018), vol.~37, Wiley Online
  Library, pp.~109--119.

\bibitem[DGK{\etalchar{*}}18]{dekel2018sparse}
\textsc{Dekel T., Gan C., Krishnan D., Liu C., Freeman W.~T.}:
\newblock Sparse, smart contours to represent and edit images.
\newblock In \emph{Proceedings of the IEEE Conference on Computer Vision and
  Pattern Recognition} (2018), pp.~3511--3520.

\bibitem[GW16]{gross2016training}
\textsc{Gross S., Wilber M.}:
\newblock Training and investigating residual nets.
\newblock \emph{Facebook AI Research} (2016).

\bibitem[HRU{\etalchar{*}}17]{heusel2017gans}
\textsc{Heusel M., Ramsauer H., Unterthiner T., Nessler B., Hochreiter S.}:
\newblock Gans trained by a two time-scale update rule converge to a local nash
  equilibrium.
\newblock In \emph{Advances in Neural Information Processing Systems} (2017),
  pp.~6626--6637.

\bibitem[HZRS16]{he2016deep}
\textsc{He K., Zhang X., Ren S., Sun J.}:
\newblock Deep residual learning for image recognition.
\newblock In \emph{Proceedings of the IEEE conference on computer vision and
  pattern recognition} (2016), pp.~770--778.

\bibitem[IZZE17]{isola2017image}
\textsc{Isola P., Zhu J.-Y., Zhou T., Efros A.~A.}:
\newblock Image-to-image translation with conditional adversarial networks.
\newblock In \emph{Proceedings of the IEEE conference on computer vision and
  pattern recognition} (2017), pp.~1125--1134.

\bibitem[JAFF16]{johnson2016perceptual}
\textsc{Johnson J., Alahi A., Fei-Fei L.}:
\newblock Perceptual losses for real-time style transfer and super-resolution.
\newblock In \emph{European conference on computer vision} (2016), Springer,
  pp.~694--711.

\bibitem[LLQ{\etalchar{*}}16]{liuLQWTcvpr16DeepFashion}
\textsc{Liu Z., Luo P., Qiu S., Wang X., Tang X.}:
\newblock Deepfashion: Powering robust clothes recognition and retrieval with
  rich annotations.
\newblock In \emph{Proceedings of IEEE Conference on Computer Vision and
  Pattern Recognition (CVPR)} (June 2016).

\bibitem[LS18]{li2018learning}
\textsc{Li Z., Snavely N.}:
\newblock Learning intrinsic image decomposition from watching the world.
\newblock In \emph{Proceedings of the IEEE Conference on Computer Vision and
  Pattern Recognition} (2018), pp.~9039--9048.

\bibitem[MLX{\etalchar{*}}17]{mao2017least}
\textsc{Mao X., Li Q., Xie H., Lau R.~Y., Wang Z., Paul~Smolley S.}:
\newblock Least squares generative adversarial networks.
\newblock In \emph{Proceedings of the IEEE International Conference on Computer
  Vision} (2017), pp.~2794--2802.

\bibitem[SZ14]{simonyan2014very}
\textsc{Simonyan K., Zisserman A.}:
\newblock Very deep convolutional networks for large-scale image recognition.
\newblock \emph{arXiv preprint arXiv:1409.1556} (2014).

\bibitem[WLZ{\etalchar{*}}18]{wang2018high}
\textsc{Wang T.-C., Liu M.-Y., Zhu J.-Y., Tao A., Kautz J., Catanzaro B.}:
\newblock High-resolution image synthesis and semantic manipulation with
  conditional gans.
\newblock In \emph{Proceedings of the IEEE conference on computer vision and
  pattern recognition} (2018), pp.~8798--8807.

\bibitem[XSA{\etalchar{*}}18]{xian2018texturegan}
\textsc{Xian W., Sangkloy P., Agrawal V., Raj A., Lu J., Fang C., Yu F., Hays
  J.}:
\newblock Texturegan: Controlling deep image synthesis with texture patches.
\newblock In \emph{Proceedings of the IEEE Conference on Computer Vision and
  Pattern Recognition} (2018), pp.~8456--8465.

\bibitem[XT15]{xie2015holistically}
\textsc{Xie S., Tu Z.}:
\newblock Holistically-nested edge detection.
\newblock In \emph{Proceedings of the IEEE international conference on computer
  vision} (2015), pp.~1395--1403.

\bibitem[ZPIE17]{zhu2017unpaired}
\textsc{Zhu J.-Y., Park T., Isola P., Efros A.~A.}:
\newblock Unpaired image-to-image translation using cycle-consistent
  adversarial networks.
\newblock In \emph{Proceedings of the IEEE international conference on computer
  vision} (2017), pp.~2223--2232.

\end{thebibliography}

% biblatex with biber
% \printbibliography                

%-------------------------------------------------------------------------

\end{document}